# Integrating selectional preferences in WordNet

**Eneko Agirre**
IXA NLP Group
University of the Basque Country
649 pk. 20.080
Donostia. Spain.
`eneko@si.ehu.es`

**David Martinez**
IXA NLP Group
University of the Basque Country
649 pk. 20.080
Donostia. Spain.
`jibmaird@si.ehu.es`

**Abstract**

Selectional preference learning methods have usually focused on word-to-class relations, e.g., a verb selects as its subject a given nominal class. This paper extends previous statistical models to class-to-class preferences, and presents a model that learns selectional preferences for classes of verbs, together with an algorithm to integrate the learned preferences in WordNet. The theoretical motivation is twofold: different senses of a verb may have different preferences, and classes of verbs may share preferences. On the practical side, class-to-class selectional preferences can be learned from untagged corpora (the same as word-to-class), they provide selectional preferences for less frequent word senses via inheritance, and more important, they allow for easy integration in WordNet. The model is trained on subject-verb and object-verb relationships extracted from a small corpus disambiguated with WordNet senses. Examples are provided illustrating that the theoretical motivations are well founded, and showing that the approach is feasible. Experimental results on a word sense disambiguation task are also provided.

## 1 Introduction

Previous literature on selectional preferences has usually learned preferences for verbs in the form of classes, e.g., the object of eat is an edible entity. This paper extends previous statistical models to classes of verbs, yielding a relation between classes in a hierarchy, as opposed to a relation between a word and a class.

The model is trained using subject-verb and object-verb associations extracted from Semcor, a corpus (Miller et al., 1993) tagged with WordNet word-senses (Miller et al., 1990), comprising around 250,000 words. The syntactic relations were extracted using the Minipar parser (Lin, 1993). A peculiarity of this exercise is the use of a sense-disambiguated corpus, in contrast to using a large corpus of ambiguous words. This corpus makes it easier to compare the selectional preferences obtained by different methods. Nevertheless, the approach can be easily applied to larger, non-disambiguated corpora.

This paper argues that class-to-class selectional preferences are a better formalization than verb-to-class models. An algorithm to integrate the acquired selectional preferences in WordNet as relations holding between synsets is provided. Some examples are described, as well as the results in a word sense disambiguation (WSD) exercise.

Following this short introduction, section 2 reviews selectional preference acquisition literature. Section 3 explains our approach, and the estimation of class frequencies is described in Section 4. Section 5 presents the algorithm for the integration in WordNet. Section 6 comments some examples of the acquired selectional preferences. Section 7 shows the results on the WSD experiment. Finally, some conclusions are drawn and future work is outlined.

## 2   Selectional preference learning

Selectional preferences try to capture the fact that linguistic elements prefer arguments of a certain semantic class, e.g. a verb like '*eat*' prefers as object edible things, and as subject animate entities, as in, (1) "*She was eating an apple*". Selectional preferences get more complex than it might seem: (2) "*The acid ate the metal*", (3) "*This car eats a lot of gas*", (4) "*We ate our savings*", etc.

Corpus-based approaches for selectional preference learning extract a number of (e.g. verb/subject) relations from large corpora and use an algorithm to generalize from the set of nouns for each verb separately. Usually, nouns are generalized using classes (concepts) from a lexical knowledge base (e.g. WordNet).

Resnik (1992, 1997) defines an information-theoretic measure of the association between a verb and nominal WordNet classes: selectional association. He uses verb-argument pairs from the Brown corpus. Evaluation is performed applying intuition and WSD. Our measure follows in part from his formalization.

Abe and Li (1996) follow a similar approach, but they employ a different information-theoretic measure (the minimum description length principle) to select the set of concepts in a hierarchy that generalize best the selectional preferences for a verb. The argument pairs are extracted from the WSJ corpus, and evaluation is performed using intuition and PP-attachment resolution.

Stetina et al. (1998) extract word-arg-word triples for all possible combinations, and use a measure of "relational probability" based on frequency and similarity. They provide an algorithm to disambiguate all words in a sentence. It is directly applied to WSD with good results.

## 3   A new approach that allows integration in WordNet

First, we will introduce the terminology used in the paper. We use concept and class indistinguishably, and they refer to the so-called *synsets* in WordNet. Concepts in WordNet are represented as sets of synonyms, e.g. *<food, nutrient>*. A word sense in WordNet is a word-concept pairing, e.g. given the concepts *a=<chicken, poulet, volaille>* and *b=<wimp, chicken, crybaby>* we can say that chicken has two word senses, the pair *chicken-a* and the pair *chicken-b*. In fact the former is sense 1 of chicken (*<chicken$_1$>*), and the later is sense 3 of chicken (*<chicken$_3$>*). For the sake of simplicity, we also say that *<chicken, poulet, volaille>* is a word sense of chicken. When a concept is taken as a class, it represents the set of concepts that are subsumed by this concept in the hierarchy.

Traditionally selectional preferences have been acquired for verbs and they do not take into account that different senses of the verbs have different preferences. Therefore, they are usually difficult to integrate in existing lexical resources as WordNet. We have extended Resnik's selectional preferences model from word-to-class (e.g. verb – nominal concepts) to class-to-class (e.g. verbal concepts – nominal concepts). The model explored in this paper emerges as a result of the following observations:

- Distinguishing verb senses can be useful. The examples for *eat* above are taken from WordNet, and each corresponds to a different word sense: example (1) is from the "*take in solid food*" sense of *eat,* (2) from the "*cause to rust*" sense, and examples (3) and (4) from the *"use up"* sense.
- If the word senses of a set of verbs are similar (e.g. word senses of ingestion verbs like *eat, devour, ingest,* etc.) they can have related selectional preferences, and we can generalize and say that a class of verbs shares the same selectional preference.

Our formalization distinguishes among verb senses; that is, we treat each verb sense as a different unit that has a particular selectional preference. From the selectional preferences of single verb senses, we also infer selectional preferences for classes of verbs. For that, we use the relation between word senses and classes in WordNet.

## 3.1 Formalization

As mentioned in the previous sections we are interested in modelling the probability of a nominal concept given that it is the subject/object of a particular verb (1) or verbal concept[1] (2):

$$P(cn_i \mid rel\, v) \qquad (1) \qquad P(cn_i \mid rel\, cv_j) \qquad (2)$$

We will now explain the three models we have tested: word-to-class, word sense-to-class (from now on referred as sense-to-class), and class-to-class. For example, we will describe the probability of the nominal concept <*chicken$_1$*> occurring as object of the verb *eat*. Examples of each of the models are provided in section 6 (cf. Table 3).

### 3.1.1 Word-to-class model: $P(cn_i \mid rel\, v)$

The probability of *eat* <*chicken$_1$*> depends on the probabilities of the concepts subsumed by and subsuming <*chicken$_1$*> being objects of eat. For instance, if *chicken$_1$* never appears as an object of eat, but other word senses under <*food, nutrient*> do, the probability of the concept <*chicken$_1$*> (first sense of chicken) will not be 0.

Formula (3) shows that for all concepts subsuming $cn_i$ the probability of $cn_i$ given the more general concept times the probability of the more general concept being a subject/object of the verb is added. The first probability is estimated dividing the class frequencies of $cn_i$ with the class frequencies of the more general concept. The second probability is estimated dividing the frequency of the general concept occurring as object of *eat* with the number of occurrences of *eat* with an object.

### 3.1.2 Sense-to-class model: $P(cn_i \mid rel\, v_j)$

Using a sense-tagged corpus, such as Semcor, we can compute the probability of the different senses of *eat* having as object the class <*chicken$_1$*>. For each sense, we use the probability formula (3) as defined in 3.1.1. In this case we have different selectional preferences for each sense of the verb: $P(cn_i \mid rel\, v_j)$.

### 3.1.3 Class-to-class model: $P(cn_i \mid rel\, cv_j)$

We compute the probability of the verb classes associated to the senses of *eat* having as object <*chicken$_1$*>, using the probabilities of all concepts above <*chicken$_1$*> being objects of all concepts above the possible senses of *eat*. For instance, if *devour* never appeared on the training corpus, the model could infer its selectional preference from that of its superclass <*ingest, take in*>.

Formula (4) shows how to calculate the probability. For each possible verb concept (*cv*) and noun concept (*cn*) subsuming the target concepts ($cn_i, cv_j$), the probability of the target concept given the subsuming concept (this is done twice, once for the verb, once for the noun) times the probability the nominal concept being subject/object of the verbal concept is added.

## 3.2 Benefits of the approach

The main benefits of our approach are the following:
- Class-to-class preferences can be trained using untagged corpora.
- In the case of sparse data the model can provide selectional preferences for word senses of verbs that do not occur in the corpus.
- Class-to-class selectional preferences can be easily integrated in WordNet.
- Distinguishes verb senses.
- Generalizes selectional preferences for classes of verbs.

---

[1] Notation: *v* stands for a verb, *cn* (*cv*) stand for nominal (verbal) concept, $cn_i$ ($cv_i$) stands for the concept linked to the *i*-th sense of the given noun (verb), *rel* could be any grammatical relation (in our case object or subject), $\subseteq$ stands for the subsumption relation, *fr* stands for frequency and $\hat{fr}$ for the estimation of the frequencies of classes.

$$P(cn_i \mid relv) = \sum_{cn \supseteq cn_i} P(cn_i \mid cn) \times P(cn \mid relv) = \sum_{cn \supseteq cn_i} \frac{\hat{fr}(cn_i, cn)}{\hat{fr}(cn)} \times \frac{\hat{fr}(cn\,relv)}{fr(relv)} \qquad (3)$$

$$P(cn_i \mid rel\,cv_j) = \sum_{cn \supseteq cn_i} \sum_{cv \supseteq cv_j} P(cn_i \mid cn) \times P(cv_j \mid cv) \times P(cn \mid rel\,cv)$$
$$= \sum_{cn \supseteq cn_i} \sum_{cv \supseteq cv_j} \frac{\hat{fr}(cn_i, cn)}{\hat{fr}(cn)} \times \frac{\hat{fr}(cv_j, cv)}{\hat{fr}(cv)} \times \frac{\hat{fr}(cn\,rel\,cv)}{fr(rel\,cv)} \qquad (4)$$

$$\hat{fr}(cn) = \sum_{cn_i \subseteq cn} \frac{1}{classes(cn_i)} \times fr(cn_i) \qquad (5)$$

$$\hat{fr}(cn_i, cn) = \begin{cases} \sum_{cn_j \subseteq cn_i} \frac{1}{classes(cn_j)} \times fr(cn_j) & \text{if } cn_i \subseteq cn \\ 0 & \text{otherwise} \end{cases} \qquad (6)$$

$$\hat{fr}(cn\,relv) = \sum_{cn_i \subseteq cn} \frac{1}{classes(cn_i)} \times fr(cn_i\,relv) \qquad (7)$$

$$\hat{fr}(cn\,rel\,cv) = \sum_{cn_i \subseteq cn} \sum_{cv_i \subseteq cn} \frac{1}{classes(cn_i)} \times \frac{1}{classes(cv_i)} \times fr(cn_i\,rel\,cv_i) \qquad (8)$$

We can keep probabilities for all possible (verbal class, nominal class) pairs, but an algorithm for pruning is also provided (cf. Section 5). Table 1 summarizes the benefits of using class-to-class selectional preferences.

|  | Learnable from: | | Apt for integration in WordNet | Distinguishes verb senses | Generalizes to classes of verbs |
|---|---|---|---|---|---|
|  | tagged corpora | untagged corpora |  |  |  |
| word to class | Yes | Yes | No | No | No |
| sense to class | Yes | No | Yes | Yes | No |
| class to class | Yes | Yes | Yes | Yes | Yes |

**Table 1:** Comparison of selectional preference models.

## 4 Estimation of class frequencies

Frequencies for classes can be counted directly from the corpus when the class is linked to a word sense that actually appears in the corpus, written as $fr(cn_i)$. Otherwise they have to be estimated using the direct counts for all subsumed concepts, written as $\hat{fr}(cn_i)$. Formula (5) shows that all the counts for the subsumed concepts ($cn_i$) are added, but divided by the number of classes for which $c_i$ is a subclass (that is, all ancestors in the hierarchy). This is necessary to guarantee the following:

$$\sum_{cn \supseteq cn_i} P(cn_i \mid cn) = 1.$$

Formula (6) shows the estimated frequency of a concept given another concept. In the case of the first concept subsuming the second it is 0, otherwise the frequency is estimated as in (5).

Formula (7) estimates the counts for [nominal-concept relation verb] triples for all possible nominal-concepts, which is based on the counts for the triples that actually occur in the corpus. All the counts for subsumed concepts are added, divided by the number of classes in order to guarantee the following:

$$\sum_{cn} P(cn \mid rel\ v) = 1$$

Finally, formula (8) extends formula (7) to [nominal-concept relation verbal-concept] in a similar way.

## 5 Algorithm for integration in WordNet.

In principle, we can take all nominal-concept and verbal-concept pairs that have a probability higher than 0 in the class-to-class model, and add a relation for each pair into WordNet. This is the model we have used for the WSD task (cf. section 8), but it adds too many relations. We have devised a pruning algorithm that chooses the highest probability nodes for each subtree combinations, discarding the rest (see Figure 1). The pruning algorithm does not affect the WSD results as it eliminates pairs that would never be selected.

We will explain the pruning step for the nominal hierarchy first. We sort all the nominal classes in the selectional preference according to their probabilities. Starting with the concept with highest probability we prune all the nominal classes whose ancestors or descendants have appeared previously in the list. This way we only take the most informative nominal class from each branch of the hierarchy. We can see an example in Figure 1. There we can observe that the concepts <b>, <c> and <e> are pruned because of the appearance of <a> higher on the table. Only <a> and <d> are left.

For a given relation *rel*
For each verb class $cv_i$ in WordNet
    Compute $P(cn_j \mid rel\ cv_i)$ for each nominal class $cn_j$ in WordNet
    Optional pruning
End For
Link all *cn,cv* pairs with *rel* if $P(cn \mid rel\ cv) > 0$

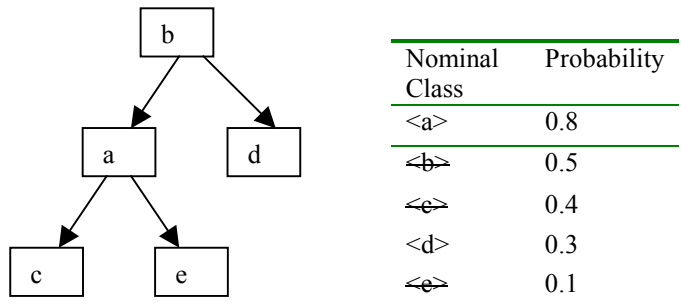

**Figure 1:** Algorithm for pruning and example of pruning for nominal classes.

This algorithm can be extended easily to pairs of nominal and verbal concepts: pairs <a,b> that have a pair <c,d> with higher probability and which either subsume both the nominal <c> and verbal <d> concepts or are subsumed by both nominal <a> and verbal <b> concepts are pruned.

## 6 Examples

We will analyze the selectional restrictions acquired for the verb *know* on the object relation. We chose this verb because it has enough occurrences in Semcor (514) and the number of senses is not too high (11). We found 87 occurrences of *know* with an object.

In Table 2 we show the nominal classes associated as object to the verb *know* using the **word-to-class** model. We only include the classes that remain after pruning, as defined in the previous section. For each class its probability is given. We see that selectional preference in the word-to-class model is confusing, and information coming from different senses (see table 3 for the list of word senses) is mixed up in the resulting table. Besides, the probability values for the different nominal classes are low.

In Table 3 we can see the results using the sense-to-class and class-to-class approaches after pruning. Because of space constraints, only the classes with highest probability for each word sense are shown. For each sense, its description in WordNet and the number of examples for which an object relation has been found in Semcor is given.

Most of the examples of *know* for which there is an object relation correspond to senses 1 and 2. Focusing on the **sense-to-class** model, we see that for sense 1 the highest probability is given to <*communication*>, which seems a good choice. Sense 2 admits a wide variety of objects, and therefore concepts that are high in the hierarchy are preferred as selectional restrictions, as <*entity, something*> or <*abstraction*>. The other senses have fewer occurrences, but anyway the system is able to detect some interesting restrictions, as <*person, individual...*> for the 4$^{th}$ sense (which gets a very high probability of 0.33) or <*idea, thought*> for the 7$^{th}$ sense.

**Class-to-class** selectional restrictions tend to use more top-level concepts and are able to provide useful information for word senses that do not appear in the corpus (e.g.: the <*make love,...*> sense of *know*). There are some differences with the preferences obtained using the sense-to-class approach, but as we will show in the next section, both representations are valid and of similar quality.

We observed that in some cases the ancestors of *know* could be very different and introduce noise, for example the ancestor <*connect, link, tie*> of the <*make love,...*> sense of *know*, but the algorithm is able to assign lower probability to those cases. To illustrate this example, in Figure 2 we show the ancestor hierarchy in WordNet for this sense of *know*, and Table 4 lists the selectional preferences for them. We can see that as the meaning of the verb ancestors gets more general, the noun concepts have lower probability. In this case, the <*make love,...*> concept assigns high probability to the correct restriction <*person, individual...* >, and the < *connect, link, tie* > concept assigns lower probabilities to its objects because the verbs belonging to this class get very different selectional preferences. In this example, the restriction <egg> gets some credit because of the incorrect identification by Minipar of the relation *mate-object-egg* from the sentence "The first worker bees do not mate or lay eggs,...".

| Nominal class | Probability |
| --- | --- |
| <person individual someone somebody mortal human soul> | 0,0778 |
| <abstraction> | 0,0656 |
| <object physical_object> | 0,0445 |
| <cognition knowledge> | 0,0380 |
| <group grouping> | 0,0330 |
| <state> | 0,0211 |
| <body_part> | 0,0202 |
| <act human_action human_activity> | 0,0190 |
| <spirit> | 0,0116 |
| <feeling> | 0,0073 |
| … | |

**Table 2:** Word-to-class selectional preferences for the objects of *know*.

| Sense | # Examples | Sense-to-class | | Class-to-class | |
|---|---|---|---|---|---|
| | | Probability | Nominal Class | Probability | Nominal Class |
| sense 1: know, cognize -- (be cognizant or aware of a fact or a specific piece of information; possess knowledge or information about; | 39 | 0,1128<br>0,0615<br>0,0535<br>0,0389<br>0,0307<br>... | <communication><br><measure quantity amount quantum><br><attribute><br><object physical_object><br><cognition knowledge><br>... | 0,1030<br>0,0386<br>0,0333<br>0,0198<br>0,0183<br>... | <abstraction><br><cognition knowledge><br><object physical_object><br><act human_action human_activity><br><group grouping><br>... |
| sense 2: know -- (know how to do or perform something; "She knows how to knit"; "Does your husband know how to cook?") | 26 | 0,0894<br>0,0507<br>0,0459<br>0,0384<br>0,0288<br>0,0172 | <entity something><br><cognition knowledge><br><abstraction><br><condition status><br><act human_action human_activity><br><wealth riches> | 0,0894<br>0,0507<br>0,0459<br>0,0384<br>0,0288<br>0,0172 | <entity something><br><cognition knowledge><br><abstraction><br><condition status><br><act human_action human_activity><br><wealth riches> |
| sense 3: know -- (be aware of the truth of something; have a belief or faith in something; regard as true beyond any doubt; "I know that I left the key on the table"; "Galileo knew that the earth moves around the sun") | 2 | 0,1509<br>0,1464 | <concept conception construct><br><Time_period period period_of_time > | 0,0952<br>0,0862<br>0,0676<br>0,0546<br>... | <abstraction><br><concept conception construct><br><person individual someone somebody mortal human soul><br><doubt uncertainty incertitude dubiety doubtfulness dubiousness><br>... |
| sense 4: know -- (be familiar or acquainted with a person or an object; "She doesn't know this composer"; "Do you know my sister?" "We know this movie") | 12 | 0,3338<br>0,1417<br>0,0286<br>0,0198 | <person individual someone somebody mortal human soul><br><group grouping><br><self ego><br><abundance copiousness> | 0,3338<br>0,1417<br>0,0286<br>0,0198 | <person individual someone somebody mortal human soul><br><group grouping><br><self ego><br><abundance copiousness> |
| sense 5: know, experience, live -- (have firsthand knowledge of states, situations, emotions, or sensations; "I know the feeling!" ) | 5 | 0,1167<br>0,0766<br>0,0621<br>0,0580 | <state><br><feeling><br><report account><br><privacy privateness seclusion> | 0,1261<br>0,1007<br>0,0933<br>... | <activity><br><cognition knowledge><br><state><br>... |
| sense 6: acknowledge, recognize, know -- (discern; "His greed knew no limits") | 2 | 0,2390<br>0,1812 | <causal_agent cause causal_agency><br><people> | 0,0776<br>0,0687<br>... | <person individual someone somebody mortal human soul><br><physical_phenomenon><br>... |
| sense 7: know -- (have fixed in the mind; "I know Latin"; | 1 | 0,2883 | <idea thought> | 0,2883 | <idea thought> |
| Sense 8: love, make out, make love, sleep with, get laid, have sex, know, do it, ... | 0 | | | 0,6798<br>0,1793<br>0,1254<br>0,0193<br>... | <person individual someone somebody mortal human soul><br><egg><br><object physical_object><br><body_part><br>... |
| .... | | | | | |

**Table 3:** Sense-to-class and class-to-class selectional preferences for the objects of *know*.

| Verb concept | Probability | Nominal class |
|---|---|---|
| <love, make out, make love, sleep with, get laid, have sex, ...> | 0.5521 | <person individual someone somebody mortal human soul> |
| <copulate, mate, pair, couple><br>make love | 0.1742<br>0.1018 | <egg><br><person individual someone somebody mortal human soul> |
| <join, conjoin><br>make contact or come together<br>"The two roads join here" | 0.0237<br>0.0152<br>0.0054<br>0.0048<br>0.0031<br>0.0028 | <person individual someone somebody mortal human soul><br><object physical_object><br><body_part><br><egg><br><dirt filth grime soil stain grease><br><communication> |
| <connect, link, tie><br>connect, fasten, or put together two or more pieces<br>"Can you connect the two loudspeakers?" | 0.0031<br>0.0019<br>0.0010<br>0.0007<br>0.0006<br>0.0004 | <object physical_object><br><person individual someone somebody mortal human soul><br><communication><br><body_part><br><happening occurrence natural_event><br><attribute> |

**Table 4:** Selectional preferences associated to the ancestor hierarchy for the <make love,...> sense of *know*.

<love, make out, make love, sleep with, get laid, have sex, ...>
   => <copulate, mate, pair, couple> -- (make love; "Birds mate in the Spring")
      => <join, conjoin> -- (make contact or come together; "The two roads join here")
         => <connect, link, tie> -- (connect, fasten, or put together two or more pieces; "Can you connect the two loudspeakers?")

**Figure 2:** Ancestor hierarchy for the <*make love,...*> sense of *know*.

## 8 Training and testing on a WSD exercise

The acquired preferences were tested on a WSD exercise. The goal is to choose the correct word sense for all nouns occurring as subjects and objects of verbs, but it could be also used to disambiguate the verbs. The method selects the word sense of the noun that is below the strongest nominal class for the verb or verb class. If more of one word sense is below the strongest class, all are selected with equal weight.

More detailed account of the experiments can be found at (Agirre and Martinez 2000; 2001). Two experiments were performed. On the lexical sample, we selected a set of 8 nouns at random and applied 10 fold cross validation to make use of all available examples. On the all-nouns experiment, we selected four files previously used in other works and tested them in turn, using the rest of the files as the training set.

Table 5 shows the data for the set of nouns of the lexical sample. Note that only 19% (15%) of the occurrences of the nouns are objects (subjects) of any verb. Table 6 shows the average results using subject and object relations for the different models. For the lexical sample task, each column shows respectively, the precision, the coverage over the occurrences with the given relation, and the recall. Random and most frequent baselines are also shown. Word-to-class gets slightly better precision than class-to-class, but class-to-class is near complete coverage and thus gets the best recall. All are well above the random baseline, but slightly below the most frequent sense (MFS).

On the all-nouns experiment, we disambiguated the nouns appearing in four files extracted from Semcor. We observed that not many nouns were related to a verb as object or subject (e.g. in the file br-a01 only 40% (16%) of the polysemous nouns were tagged as object (subject)). Table 6 shows the average recall obtained on this task. Class-to-class attains the best recall.

We think that given the small corpus available, the results are good. Note that there is no smoothing or cut-off value involved, and some decisions are taken with very little points of data. Sure enough, both smoothing and cut-off values will permit to improve the precision. On the contrary, literature has shown that the most frequent sense baseline needs less training data.

| Noun | # sens | # occ | # occ. as obj | # occ. as subj |
|---|---|---|---|---|
| account | 10 | 27 | 8 | 3 |
| age | 5 | 104 | 10 | 9 |
| church | 3 | 128 | 19 | 10 |
| duty | 3 | 25 | 8 | 1 |
| head | 30 | 179 | 58 | 16 |
| interest | 7 | 140 | 31 | 13 |
| member | 5 | 74 | 13 | 11 |
| people | 4 | 282 | 41 | 83 |
| Overall | 67 | 959 | 188 | 146 |

**Table 5**: Data for the selected nouns.

| | Lexical sample (8 nouns) | | | | | | All-nouns (4 files) | |
|---|---|---|---|---|---|---|---|---|
| | Obj. | | | Subj. | | | Obj. | Subj. |
| | Prec. | Cov. | Rec. | Prec. | Cov. | Rec. | Rec. | Rec. |
| Random | .192 | 1.00 | .192 | .192 | 1.00 | .192 | .265 | .296 |
| MFS | .690 | 1.00 | .690 | .690 | 1.00 | .690 | .698 | .790 |
| Word2class | .669 | .867 | .580 | .698 | .794 | .554 | .424 | .597 |
| Class2class | .666 | .973 | .648 | .680 | .986 | .670 | .506 | .692 |

**Table 6**: Average results for the 8 nouns and the 4 files.

# 9 Conclusions

We presented a statistical model that extends selectional preference to classes of verbs, yielding a relation between classes in a hierarchy, as opposed to a relation between a word and a class. The motivation to depart from word-to-class models is twofold: different senses of a verb may have different preferences, and some classes of verbs can share preferences. Besides, the model can be trained on untagged corpora, and has the following advantages over word-to-class models: it can provide selectional preferences for word senses of verbs that do not occur in the corpus via inheritance, and it provides a model that can be integrated easily as a relation between concepts in WordNet. In this sense, an algorithm to integrate the acquired class-to-class selectional restrictions in WordNet in a sensible way has also been described.

The model is trained using subject-verb and object-verb relations extracted from a sense-disambiguated corpus by Minipar. A peculiarity of this exercise is the use of a sense-disambiguated corpus, in contrast to using a large corpus of ambiguous words. Evaluation is based on a word sense disambiguation exercise for a sample of words and a sample of documents from Semcor. The proposed model gets similar results on precision but significantly better recall than the classical word-to-class model. This can be explained by the fact that the class-to-class model generalizes well and is able to provide sensible selectional preferences to verb senses not seen in the training data.

For the future, we plan to train the model on a large untagged corpus, in order to compare the quality of the acquired selectional preferences with those extracted from this small tagged corpora. The model can easily be extended to disambiguate other relations and PoS, and we plan to measure the effectiveness for other PoS.